%% file: root.tex
\documentclass[letterpaper, 10 pt, conference]{ieeeconf}  % Comment this line out if you need a4paper

\IEEEoverridecommandlockouts                              % This command is only needed if 
\usepackage{hyperref}
\hypersetup{bookmarksopen,bookmarksnumbered,
pdfpagemode=UseOutlines,
colorlinks=true,
linkcolor=blue,
anchorcolor=blue,
citecolor=blue,
filecolor=blue,
menucolor=blue,
urlcolor=blue
}
\usepackage{upgreek}
\usepackage{amsfonts,amssymb}
\usepackage{bm}
\usepackage{bbm}

\usepackage{amsthm} % for new theorem
\usepackage{thmtools}
\usepackage{mathtools}
\usepackage{xspace}
\usepackage{units}
\usepackage{booktabs} %for table rulers
\usepackage{tabulary}
\usepackage[usenames,dvipsnames,table]{xcolor} %Options expand the named colours
\usepackage{diagbox}
\usepackage[ruled,vlined,linesnumbered]{algorithm2e}  %updated version of algorithm2e.sty is in ./ folder.
\usepackage{parskip}
\SetKwComment{Comment}{$\triangleright$\ }{}
\usepackage{ifthen,version}
\usepackage{soul}
\usepackage{csquotes}
\usepackage{microtype}      % microtypography
\usepackage{lipsum}
\usepackage{tikz}
\let\labelindent\relax
\usepackage{todonotes}
\usepackage{enumitem}
\usepackage[noadjust]{cite}
\usepackage{wrapfig}
\usepackage{graphicx}
\usepackage{array}

\usepackage[normalem]{ulem}
\newcommand\redsout{\bgroup\markoverwith{\textcolor{red}{\rule[0.5ex]{2pt}{0.7pt}}}\ULon}

\newcommand{\fullFigGap}[0]{\vspace{-1.5\baselineskip}} %Use this command at the end of the caption of a full-page figureto eat up some unnecessary whitespace.

\newcommand{\xxnote}[3]{}
\ifx\hidenotes\undefined
  \usepackage{color}
  \renewcommand{\xxnote}[3]{\color{#2}{#1: #3}}
\fi

\usepackage{multirow}
\usepackage{pbox}

\newtheoremstyle{hypstyle}
{3pt} % Space above
{3pt} % Space below
{\itshape} % Body font
{} % Indent amount
{\bfseries} % Theorem head font
{.} % Punctuation after theorem head
{.5em} % Space after theorem head
{} % Theorem head spec (can be left empty, meaning `normal')

\theoremstyle{hypstyle}

\input{macros/math_operators}

\input{macros/math_defns}
\input{macros/text_shortcuts}
\graphicspath{{figs/}}

\title{\LARGE \bf
Learning Visuomotor Policies for Aerial Navigation\\ Using Cross-Modal Representations %\rbnote{is this better?}
\vspace{-2mm}}
\author{Rogerio Bonatti$^{*1}$, Ratnesh Madaan$^{2}$, Vibhav Vineet$^{2}$, Sebastian Scherer$^{1}$, and Ashish Kapoor$^{2}$ 
\vspace{-1.36mm}% <-this % stops a space
\thanks{$^{1}$The Robotics Institute, Carnegie Mellon University, Pittsburgh PA
        {\tt\small \{rbonatti, basti\}@cs.cmu.edu}}%
\thanks{$^{2}$Microsoft Corporation, Redmond, WA
        {\tt\small \{ratnesh.madaan, vibhav.vineet, akapoor\}@microsoft.com}}%
\thanks{$^{*}$ Work done while interning at Microsoft Corporation, Redmond}
}

\begin{document}

\maketitle
\thispagestyle{empty}
\pagestyle{empty}

% %%%%%%%%%%%%%%%%%%%%%%%%%%%%%%%%%%%%%%%%%%%%%%%%%%%%%%%%%%%%%%%%%%%%%%%%%%%%%%%

\input{inputs/0_abstract}
\input{inputs/1_intro}
\input{inputs/2_related_work}
\input{inputs/3_approach}
\input{inputs/4_exp}

\input{inputs/5_discussion}

\section*{ACKNOWLEDGMENTS}

We thank Matthew Brown, Nicholas Gyde, Christoph Endner, Lorenz Stangier, and Nicolas Iskos for the help with experiments, and Debadeepta Dey for insightful discussions.

%%%%%%%%%%%%%%%%%%%%%%%%%%%%%%%%%%%%%%%%%%%%%%%%%%%%%%%%%%%%%%%%%%%%%%%%%%%%%%%%

% \bibliographystyle{IEEEtran}
% \bibliography{IEEEabrv,IEEEexample}
\footnotesize{
\bibliographystyle{IEEEtran}
\bibliography{root.bbl}
}

% \bibliographystyle{IEEEtran}
% {\small
%   \bibliography{IEEEexample}  % .bib
% }

\end{document}

%% file: macros/math_operators.tex
\DeclareMathOperator*{\argmin}{arg\,min}

\newcommand{\argminprob}[1]{\underset{#1}{\argmin}}

\newcommand{\expect}[2]{\mathbb{E}_{#1}\left[#2\right]}

\newcommand{\real}[0]{\mathbb{R}}

\newcommand{\bbm}{\begin{bmatrix}}
\newcommand{\ebm}{\end{bmatrix}}

%% file: inputs/0_abstract.tex
% !TEX root = ../root.tex

% \vspace{-2mm}
\begin{abstract}
Machines are a long way from robustly solving open-world perception-control tasks, such as first-person view (FPV) aerial navigation. While recent advances in end-to-end Machine Learning, especially Imitation and Reinforcement Learning appear promising, they are constrained by the need of large amounts of difficult-to-collect labeled real-world data. Simulated data, on the other hand, is easy to generate, but generally does not render safe behaviors in diverse real-life scenarios.
%\redsout{In this work we propose to learn rich representations and policies by leveraging unsupervised data, such as video footage from an FPV drone, together with easy to generate simulated labeled data.}
In this work we propose a novel method for learning robust visuomotor policies for real-world deployment which can be trained purely with simulated data.
We develop rich state representations that combine supervised and unsupervised environment data. 
Our approach takes a cross-modal perspective, where separate modalities correspond to the raw camera data and the system states relevant to the task, such as the relative pose of gates to the drone in the case of drone racing. 
We feed both data modalities into a novel factored architecture, which learns a joint low-dimensional embedding via Variational Auto Encoders. This compact representation is then fed into a control policy, which we trained using imitation learning with expert trajectories in a simulator.
% \redsout{Such joint representations allow us to leverage rich labeled information from simulations together with the diversity of possible experiences via the unsupervised real-world data.} 
% \rbnote{do we still need this sentence saying that we can use unsupervised real-world data given that it did not improve experiments? probably remove since this is not a major finding that should be in abstract}
% We present experiments to analyze the rich latent spaces learned with our proposed representations. 
We analyze the rich latent spaces learned with our proposed representations, and
% In simulation, we 
show that the use of our cross-modal architecture significantly improves control policy performance as compared to end-to-end learning or purely unsupervised feature extractors.
We also present real-world results for drone navigation through gates in different track configurations and environmental conditions. Our proposed method, which runs fully onboard, can successfully generalize the learned representations and policies across simulation and reality, significantly outperforming baseline approaches.

Supplementary video: \url{https://youtu.be/VKc3A5HlUU8}\\
Open-sourced code available at: \small{\url{https://github.com/microsoft/AirSim-Drone-Racing-VAE-Imitation}}

% \rbnote{re-write to convey the three main points that we have in this paper, and to reflect our experiments}

% \rbnote{a few things to consider for abstract: - Context: What is the context of your work, what is the start of the art 
% - Need: What is the lack of the start of the art 
% - Task: What is the task you want to address in your work (should be 1 sentence) 
% - Object: How do you plan to address your task? 
% - Results: 
% - Conclusions: What are your expected conclusions?}

%\rbnote{write after we're done with the other parts of the paper}

\end{abstract}

%% file: inputs/1_intro.tex
% !TEX root = ../root.tex

%%%%%%%%%%%%%%%%%%%%%%%%%%%%%%%%%%%%%%%%%%%%%%%%%%%%%%%%%%%%%%%%%%%%%%%%%%%%%%%%
\vspace{-2mm}
\section{Introduction}
\vspace{-2mm}
%First-person view (FPV) racing of drones is an exemplary feat of human mind. 
Aerial navigation of drones using first-person view (FPV) images is an exemplary feat of the human mind. 
Expert pilots are able to plan and control a quadrotor with high agility using a potentially noisy monocular camera feed, without comprising safety. 
We are interested in exploring the question of what would it take to build autonomous systems that achieve similar performance levels. 
%This is a difficult problem due to multiple reasons. %\rbnote{these first sentences are quite focused on drone racing, but I think they're good. I wouldn't change it}
%First, the high dimensional nature and drastic variability of the input image data requires robust representations invariant to visual appearance and artifacts.
%Second, the image-to-action drone navigation task is a non-myopic sequential decision-making problem, and popular end-to-end control methods such as end-to-end reinforcement and even imitation learning are sample inefficient and/or not robust to perceptual changes in the images \cite{kang2019generalization,tai2017virtual,Zhu_MarrNet_abs18}. 
%Lastly, end-to-end training of real-world systems is highly impractical, especially in the initial phases of training when the learnt policy is highly prone to errors and collisions.

% \vspace{-0.5mm}
\vspace{-0.2mm}

One of the biggest challenges in the navigation task is the high dimensional nature and drastic variability of the input image data. Successfully solving the task requires a representation that is invariant to visual appearance and robust to the differences between simulation and reality. Collecting labeled real-world data to minimize the sim-to-real gap, albeit possible, requires intensive effort and specialized equipment for gathering ground-truth labels \cite{kaufmann2018deep,kaufmann2019beauty}. 
Attempting to solve the task solely with real-world data is also challenging due to poor sample efficiency of end-to-end methods, and often leads to policies that are unable to deal with large perceptual variability \cite{kang2019generalization,tai2017virtual,Zhu_MarrNet_abs18}. Additionally, end-to-end training in the real world is expensive and dangerous, especially in early phases of training when policies are highly prone to errors and collisions.

%\rbnote{re-write these 3 main challenges, the language is a bit convoluted, specially for the second one. 1. we want representation invariance btw sim-real visual appearances -- therefore we might try to train in real life, but current techniques are 2. sample inefficient, and therefore 3. doing this in a real robot is dangerous / expensive / impractical}.

% \rmnote{previous para could use some citations. perhaps factor in the existing drone racing papers}

\vspace{-0.2mm}

Sim-to-real transfer learning methods aim to partially alleviate these challenges by training policies in a synthetic environment and then deploying them in the real-world \cite{sadeghi2016cad,ganin2016domain,tobin2017domain}. 
%A key challenge here is that the policy needs to be resilient to appearance and physics  differences between simulation and the reality, which is often handled by domain randomization \cite{wu2018building,loquercio2019deep}. 
Domain randomization uses a large collection of object appearances and shapes, assuming that any real-life features encountered will be represented within a subset of the database.
Simulations can be used to generate large amounts of synthetic data under a wide variety of conditions \cite{wu2018building,loquercio2019deep}.
% Collecting labeled real-world data to minimize the sim-to-real gap, albeit possible, requires intensive effort and specialized equipment for gathering ground-truth labels \cite{kaufmann2018deep,kaufmann2019beauty}.% for drone racing.
% For example, real-world data collection for drone racing requires significant instrumentation of the environment and the robot, and is often impractical.

% collecting real-world data for learning is difficult as there is no easy way to get the 
% Often the key to achieving such success is via introducing some real-world data which then is augmented with the simulated data to train the system.  

\begin{figure}[t]
    \centering
    \includegraphics[width=0.49\textwidth]{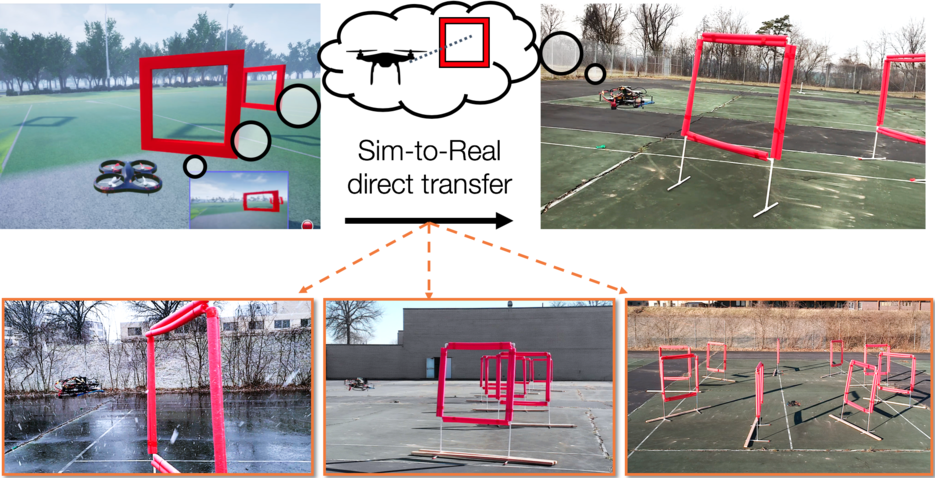}
    \vspace{-6mm}
    \caption{The proposed framework uses simulations to learn a rich low-dimensional state representation using multiple data modalities. This latent vector is used to learn a control policy which directly transfers to real-world environments. We successfully deploy the system under various track shapes and weather conditions, ranging from sunny days to strong snow and wind.}
    \label{fig:main_iros}
    \vspace{-5.5mm}
\end{figure}

% \rmnote{we might want to talk/cite about scaramuzza and/or shaojie drone racing papers in intro}

%\redsout{A key observation in this work is that unsupervised data is relatively easy to gather in real environments. 
%Consequently, a learning framework that can leverage unsupervised first-person-view videos can have a significant advantage. 
%In this paper, we explore the question of combining such unsupervised data which is readily available in real-life, along with labeled simulated data which can be easily generated with high-fidelity simulators \cite{airsim2017fsr}.
%Unsupervised data helps in fine-tuning policies trained only on simulated data, thus helping us bridge the simulation-reality gap.} 
%\rbnote{need to re-write this into a new key observation. We should change the key observation to the fact that the introduction of the cross-modality in learning a representation can serve as a regularization term to the model. therefore, feature extraction won't overfit to the specificities of simulation data. This generates good sim2real transfer, and is more efficient and principled than domain randomization, where one can only hope that the real life appearances are contained within the subset of varying shapes and textures -- not sure if this last claim is too strong}

\vspace{-0.2mm}

In this work, we introduce cross-modal learning for generating representations which are robust to the simulation-reality gap, and do not overfit to specificities of the simulated data.
In particular, the need to explain multiple modes of information during the training phase aids in implicit regularization of the representation, leading to an effective transfer of models trained in simulation into the real world.
% Furthermore, our proposed framework also has the ability to make use of unsupervised data in the training phase, thereby allowing us to incorporate real-life unlabeled traces into the state representation. 
We use high-fidelity simulators \cite{airsim2017fsr} to generate both labeled and unlabeled modalities of simulated data.
% , to train policies that bridge the simulation-reality gap. 
Furthermore, our proposed framework can also use unlabeled real-world data in the training phase, thereby allowing us to incorporate real-life unlabeled traces into the state representation.
Fig.~\ref{fig:main_iros} depicts the overall concept, showing a single perception module shared for simulated and real autonomous navigation.
% Fig.~\ref{fig:main_iros} depicts the overall framework, showing the decomposition of the autonomous navigation problem into the tasks of representation and control policy learning.
Our unmanned aerial vehicle (UAV) uses FPV images to extract a low-dimensional representation, which is then used as the input to a control policy network to determine the next control actions. We successfully test the system operating in challenging conditions, many of which were previously unseen by the simulator, such as snow and strong winds.

%\rbnote{overall this paragraph is a bit confusing, and we need to re-write -- there are two key ideas we need to transmit: 1. that we decide to break down the end-to-end process into learning a representation and learning a control policy, and 2. that cross-modality acts as a regularizer, dramatically improving sim2real transfer without the need for the "open-loop" process of domain randomization. }

% For drone racing not only it is easy to obtain unsupervised date in the form of FPV videos shot from a camera mounted in the drone, but there are  that can generate quality supervised data.
%While there has been work in training a separate world representation using unsupervised learning \cite{ha2018world}, however, there is no way to enforce that this representation captures the features that are actually meaningful for the task in hand. 

\vspace{-0.2mm}

Our proposed framework entails learning a cross-modal representation for state encoding. 
The first data modality considers the raw unlabeled sensor input (FPV images), while the second directly characterizes state information directly relevant for the desired application.
In the case of drone racing, our labels correspond to the relative pose of the next gate defined in the drone's frame.
% The key idea is to learn a joint low-dimensional representation across the modalities that encode information both about the sensor observations (e.g. images) and the system state (e.g. pose of drone with respect to a gate). 
We learn a low-dimensional latent representation by extending the Cross-Modal Variational Auto Encoder (CM-VAE) framework from \cite{spurr2018cross}, which uses an encoder-decoder pair for each data modality, while constricting all inputs and outputs to and from a single latent space.
Consequently, we can naturally incorporate both labeled and unlabeled data modalities into the training process of the latent variable.
% Consequently, the latent variable can be encoded from both modalities, allowing us to incorporate both the labeled and the unlabeled data naturally into the training process.
% The latent low-dimensional representation is modeled 
% via extensions to  cross-modal Variational Auto Encoder (VAE) framework \cite{} that explicitly models projections and reconstructions to and from the joint space across the modalities. 
%In addition, this framework forces the image encoder to compress the latent variable taking into account features relevant to both data modalities. \rmnote{is previous sentence redundant?}
% The learnt parameters of the network enables us to recover the low-dimensional representation summarizing information across the modalities. 
We then use imitation learning to train a control policy that maps latent variables into velocity commands for the UAV. 
The representation uses only raw images during deployment, without access to the ground-truth gate poses.
% using imitation or reinforcement learning 
% that operates over this representation, 
% rendering robust sim-to-real domain transfer.
While closest to our work is the recent line of research on autonomous drone racing \cite{kaufmann2018deep,loquercio2019deep}, we would like to note that our objective here is not to engineer a system that necessarily finishes the race the fastest.
Unlike the prior work, we do not assume prior knowledge during deployment in terms of an accurate dynamics model coupled with optimal trajectory generation methods. 
Instead, our goal is to learn visuomotor policies operating on learned representations that can be transferred from simulation to reality. Thus, the methods presented in this paper are not directly comparable to \cite{kaufmann2018deep,loquercio2019deep}.

While in this paper we specifically focus on the problem of aerial navigation in a drone racing setting, the proposed techniques are general and can be applied to other perception-control tasks in robotics. Our key contributions are:
\begin{itemize}
\item We present a cross-modal framework for learning latent state representations for navigation policies that use unsupervised and supervised data, and interpret the bi-modal properties of the latent space encoding;
\item We provide simulation experiments comparing variants of the cross-modal framework with baseline feature extractors such as variational auto-encoders (VAEs), gate pose regression, and end-to-end learning;
\item We provide multiple real-world navigation experiments and performance evaluations of our control policies. We show that our proposed representation allows for sim-to-real deployment of models learned purely in simulation, achieving over one kilometer of cumulative autonomous flight through obstacles.
\end{itemize}

%% file: inputs/2_related_work.tex
\section{Related Work} 
\label{sec:related_work}

% \rbnote{refer to this folder with examples of papers for lit review:
% \tiny{https://drive.google.com/drive/folders/1ifG4lIrv3qGFh4AXxzBJQHpSMRcl5CaB?usp=sharing}
% }

% There is a large body of literature on navigation policies, representation learning and drone racing.
% We briefly discuss some of the most relevant works: 
\paragraph{Navigation policies}
Classically, navigation policies rely on state estimation modules that use either visual-inertial based odometry \cite{delmerico2018benchmark} or simultaneous localization and mapping \cite{sturm2012benchmark}. 
These techniques can present high drift and noise in typical field conditions, impacting the quality of both the robot localization and the map representation used for planning. Therefore, trajectory optimization based algorithms \cite{mellinger2011minimum, oleynikova2016continuous,bonatti2019towards} can result in crashes and unsafe robot behaviors.
Against these effects, \cite{Ross_2013_7410} learn a collision avoidance policy in dense forests using only monocular cameras, and \cite{loquercio2018dronet} learn a steering function for aerial vehicles in unstructured urban environments using driving datasets for supervision. 
% \rbnote{mention a couple more papers on navigation for robots / UAVs?}

% <<<<<<< HEAD
% A rich body of work focuses on developing navigation policies that can operate reactively even under degraded state estimation, combining the perception and control tasks. For example, \cite{Ross_2013_7410} learned to control an aerial robot to avoid collisions in dense forests using only monocular cameras. Similarly, \cite{loquercio2018dronet} learned a steering function for aerial vehicles in unstructured urban environments using driving datasets for supervision. \rbnote{mention a couple more papers on navigation for robots / UAVs?}

% Recent work from machine learning community explored learning separate networks for the environment representation and controls \cite{ha2018world,lesort2018state,hafner2018learning}, instead of the end-to-end paradigm. The goal of intermediate representations is to learn a (low-dimensional) semantically invariant latent codes. Such intermediate representations means behavior cloning or reinforcement learning methods search over a smaller policy class \cite{lesort2018state}, thus requiring less data and helping to overcome the curse of dimensionality.
% =======

Recently, \cite{ha2018world,lesort2018state,hafner2018learning,kang2019generalization} explore learning separate networks for the environment representation and controls, instead of the end-to-end paradigm. The goal of an intermediate representations is to extract a low-dimensional space which summarizes the key geometrical properties of the environment, while being invariant to textures, shapes, visual artifacts. 
Such intermediate representations mean that behavior cloning or reinforcement learning methods have a smaller search space \cite{lesort2018state} and more sample efficiency. 
% \rbnote{talk about \cite{kang2019generalization} as well, and how they break down navigation into perception learned in sim and navigation learned in real}

\paragraph{Learning representations for vision}
Variational Autoencoder (VAE) based approaches have been shown to be effective in extracting low-dimensional representation from image data \cite{Kingma_VAE_ICLR14, Higgins_betavae_ICLR17, Kim_factorvae_arXiv18, Burgess_BMONet_CoRR19}.
% These are probabilistic encoder-decoder architectures where the encoders learn low-dimensional latent parameters to represent the data. In general, both the encoders and decoders are deep networks. 
Recently, VAEs have been to leveraged to extract representations from multiple modalities \cite{aytar2017cross, ngiam2011multimodal, liong2017cross, spurr2018cross}. 
Relevant to our work, \cite{spurr2018cross} propose a cross-modal VAE to learn a latent space that jointly encodes different data modalities (images and 3D kyepoints associated with hand joints) for a image to hand pose estimation problem. 

%Next we briefly describe current state-of-the-art methods for drone racing. 
\paragraph{Drone Racing}
% \rbnote{keep this as drone racing or drone navigation, as more general? I think racing is fine as long as we have the paragraph before in intro}
% \rbnote{or should the comparison paragraph be here instead of intro?}
We find different problem definitions in the context of autonomous drone racing. 
\cite{loquercio2019deep,kaufmann2018deep} focus on scenarios with dynamic gates by decoupling perception and control.
They learn to regress to a desired position using monocular images with a CNN, and plan and track a minimum jerk trajectory using classical methods  \cite{mellinger2011minimum, richter2016polynomial}.
%, where desired values are obtained using an optimal trajectory \cite{mellinger2011minimum}. 
% During  then achieved using classical trajectory fitting and following methods. 
\cite{loquercio2019deep} utilize domain randomization for effective simulation to reality transfer of learned policies. 
% One downside of such methods is the need for large track-specific datasets to learn good quality behaviors.

Gate poses are assumed as \textit{a priori} unknown in \cite{jung2018direct, jung2018perception,jung2018real}.
\cite{jung2018direct} use depth information and a guidance control law for navigation.
\cite{jung2018perception,jung2018real} use a neural network for gate detection on the image. 
A limitation of the guidance law approach is that the gate must be in view at all times, and it does not take into account gate relative angles during the approach. 

\cite{kaufmann2019beauty} formulate drone racing as flight though a predefined ordered set of gates. 
They initialize gate locations with a strong prior via manual demonstration flights. 
The belief over each gate location is then updated online by using a Kalman Filter over the gate pose predictions from a neural network.

% Our work addresses the limitations of previous works. 
In our work we take inspirations from the fields of policy and representation learning to present a method that combines unsupervised and supervised simulated data to train a single latent space on which a control policy acts. 
The bi-modal nature of the latent space implicitly regularizes the representation model, allowing for policy generalization across the simulation-reality gap.

%% file: inputs/3_approach.tex
\section{Approach} 
\label{sec:approach}

\begin{figure}[t]
    \centering
    \includegraphics[width=0.44\textwidth]{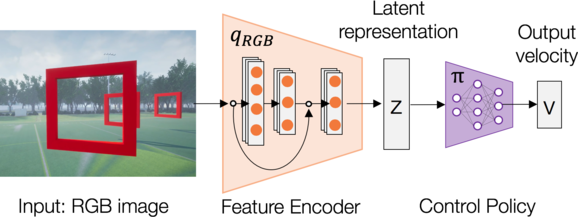}
    \caption{Control system architecture. The input image is encoded into a latent representation of the environment. A control policy acts on the lower-dimensional embedding to output the desired robot velocity commands.}
    \label{fig:process}
\end{figure}

This work addresses the problem of robust autonomous navigation through a set of gates with unknown locations. 
Our approach is composed of two steps: first, learning a latent state representation, and second, learning a control policy operating on this latent representation (Fig.~\ref{fig:process}). 
The first component receives monocular camera images as input and encodes the relative pose of the next visible gate along with background features into a low-dimensional latent representation. 
This latent representation is then fed into a control network, which outputs a velocity command, later translated into actuator commands by the UAV's flight controller.
% , which translates them into actuator commands. 
 % depicts the process.

\subsection{Definitions and Notations}
Let $W$ define the world frame, $B$ the body frame, and $G_i$ the frame of the target gate. 
Let $E$ define the full environment geometry and object categories. 
Assuming that all gates are upright, let $y_i = [r, \theta, \phi, \psi]$ define the relative spherical coordinates and yaw gate frame $G_i$, in the $B$ frame.
% $\{r, \theta, \phi \}$ define translation $t_{BG_i}$ in polar coordinates, and $\psi$ represents the relative yaw angle.

We define $q_{RGB}(I_t) \rightarrow \real^N$ to be an encoder function that maps the current image $I_t$ to a latent compressed vector $z_t$ of size $N$. 
Let $\pi(z_t) \rightarrow \real^4$ be a control policy that maps the current encoded state to a body velocity command $v_B = [v_x, v_y, v_z, v_{\psi}]$, corresponding to linear and yaw velocities.

Let $\pi^*$ be an expert control policy. 
% in contrast to our policy's partial state information $q_{RGB}(I_t)$. 
Our objective is to find the optimal model parameters $\Theta^*$ and $\Phi^*$ that minimize the expectation of distance $D$ between our control policy and the expert, taken over observed states $s$. Note that the expert policy operates with full knowledge of the environment $E$, while our policy only has access to the observation $q_{RGB}(I_t)$:
\begin{equation}
\begin{aligned}
\label{eq:main_obj}
% \theta^*, \Phi^* &= \argminprob{\theta, \Phi} \quad D\{\pi^*, \pi(q(I))\}
\Theta^*, \Phi^* &= \argminprob{\Theta, \Phi} \quad \expect{s}{D \Big( \pi^* \big( E \big), \ \pi^\Phi \big( q^\Theta_{RGB}(I \big) \Big)}
\end{aligned}
\end{equation}
% \rmnote{you need expectation over environment dynamics in the above loss function? See the loss function in mohak's SAIL paper. as you have to roll out the policy.}

\subsection{Learning Cross-Modal Representations for Perception}
\label{subsec:app_rep}

The goal of the perception module is to extract all pertinent information for the current task from the current state of the environment $E$ and UAV. Several approaches exist for feature extraction, from fully supervised to fully unsupervised methods, as mentioned in Section~\ref{sec:related_work}. 

An effective dimensionality reduction technique should be smooth, continuous and consistent \cite{spurr2018cross}, and in our application, also be robust to differences in visual information across both simulated and real images. To achieve these objectives we build on the architecture developed by \cite{spurr2018cross}, who use a cross-modal variant of variational auto-encoders (CM-VAE) to train a single latent space using multiple sources of data representation.
The core hypothesis behind our choice of encoding architecture is that, by combining different data modalities into one latent vector, we can induce a regularization effect to prevents overfitting to one particular data distribution.
As shown in Section~\ref{subsec:exp_real}, this becomes an important feature when we transfer a model learned with simulated data to the real world, where appearances, textures and shapes can vary significantly.
% \rbnote{review this sentence later. maybe there's a better way to express it.}

% \rbnote{should I talk here about the regularization role that the cross-modalities have on the real-life output? at this point of the paper it might be a bit out of context. probably better to talk about that during the real-life results when the comparison is clear}

\begin{figure}[t]
    \centering
    \includegraphics[width=0.42\textwidth]{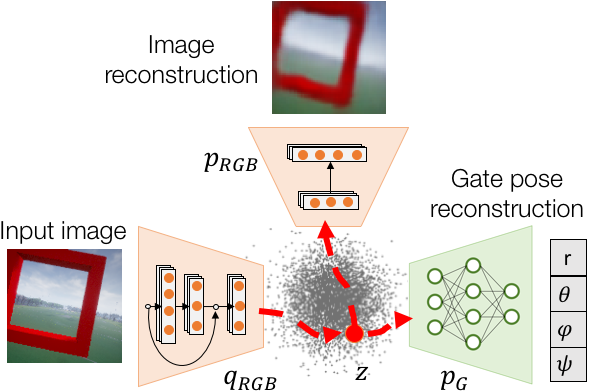}
    \caption{Cross-modal VAE architecture. Each data sample is encoded into a single latent space that can be decoded back into images, or transformed into another data modality such as the poses of gates relative to the UAV.}
    \label{fig:arch}
\end{figure}

\noindent{\it CM-VAE derivation and architecture:}
The cross-modal architecture works by processing a data sample $x$, which can come from different modalities, into the same latent space location (Fig.~\ref{fig:arch}). In robotics, common data modalities found are RGB or depth images, LiDAR or stereo pointclouds, or 3D poses of objects in the environment. 
In the context of drone racing, we define data modalities as RGB images and the relative pose of the next gate to the current aircraft frame, \textit{i.e}, $x_{RGB} = I_t$ and $x_{G} = y_i = [r, \theta, \phi, \psi]$. The input RGB data is processed by encoder $q_{RGB}$ into a normal distribution $\mathcal{N}(\mu_t, \sigma_t^2)$ from which $z_t$ is sampled. Either data modality can be recovered from the latent space using decoders $p_{RGB}$ and $p_{G}$.

In the standard definition of VAEs, the objective is to optimize the variational lower bound on the log-likelihood of the data \cite{rezende2014stochastic,Kingma_VAE_ICLR14}. In \cite{spurr2018cross}, this loss is re-derived to account for probabilities across data modalities $x_i$ and $x_j$, resulting in the new lower bound shown in Eq.~\ref{eq:bound}: 

\begin{equation}
\begin{aligned}
\label{eq:bound}
\expect{z \sim q(z|x_i)}{\log p(x_j|z)} - D_{KL}(q(z|x_i) || p(z))
\end{aligned}
\end{equation}

We use the Dronet \cite{loquercio2018dronet} architecture for encoder $p_{RGB}$, which is equivalent to an 8-layer Resnet \cite{he2016deep}. We choose a small network, with about $300$K parameters, for its low onboard inference time. For the image decoder $q_{RGB}$ we use six transpose convolutional layers, and for the gate decoder $p_{G}$ we use two dense layers.

\noindent{\it Training procedure:}
We follow the training procedure outlined in Algorithm~1 of \cite{spurr2018cross}, considering three losses: 
(i) MSE loss between actual and reconstructed images $(I_t, \hat{I_t})$, (ii) MSE loss for gate pose reconstruction $(y_i, \hat{y_i})$, and (iii) Kullback-Leibler (KL) divergence loss for each sample. 
% $L_{i} = \abs{x_i-\hat{x_i}}_2$. 
During training, for each unsupervised data sample we update networks $q_{RGB}$, $p_{RGB}$, and for each supervised sample we update both image encoder $q_{RGB}$ and gate pose decoder $p_{G}$ with the gradients. 

\begin{figure*}[t]
    \centering
    \includegraphics[width=0.94\textwidth]{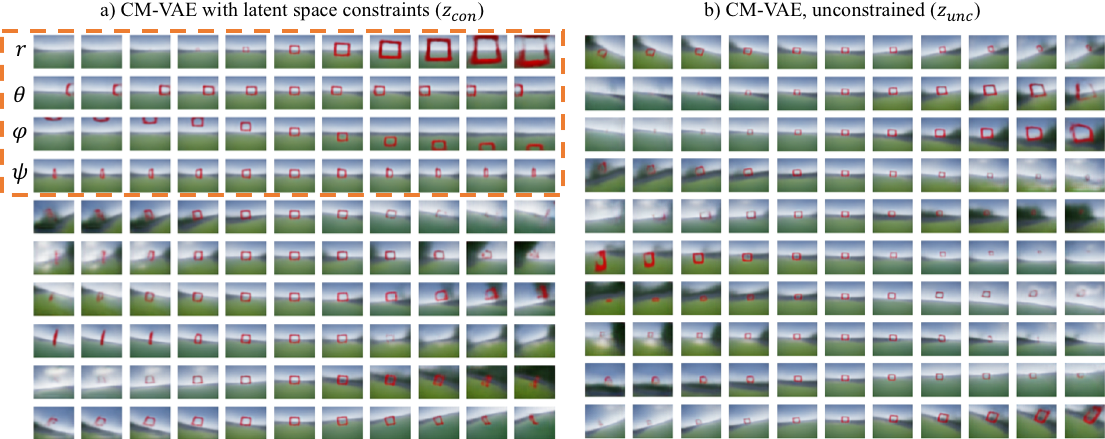}
    \caption{Visualization of latent space from a) constrained and b) unconstrained cross-modal representations. The constraints on latent space force the disantanglement of the first four variables of $z_{con}$ to encode the relative gate pose, condition that is also observed in the image modality.}
    \label{fig:z_vis}
\end{figure*}

\noindent{\it Imposing constraints on the latent space:}
Following recent work in distantangled representations \cite{Higgins_betavae_ICLR17,kim2018disentangling}, we compare two architectures for the latent space structure. Our goal is to improve performance of the control policy and interpretability of results. In first architecture, $z_{unc}$ stands for the unconstrained version of the latent space, where: $\hat{y_i} = p_{G}(z_{unc})$ and $\hat{I_t} = p_{RGB}(z_{unc})$. 
For second architecture, instead of a single gate pose decoder $p_{G}$, we employ 4 independent decoders for each gate pose component, using the first 4 elements of $z_{con}$. As human designers, we know that these features are independent (\textit{e.g}, the distance between gate and drone should have no effect on the gate's orientation).
Therefore, we apply the following constraints to $z_{con}$: $\hat{r} = p_{r}(z_{con}^{[0]})$, $\hat{\theta} = p_{\theta}(z_{con}^{[1]})$, $\hat{\psi} = p_{\psi}(z_{con}^{[2]})$, $\hat{\phi} = p_{\phi}(z_{con}^{[3]})$. The image reconstruction step still relies on the full latent variable: $\hat{I_t} = p_{RGB}(z_{con})$.

% \rbnote{

% Describe the hypothesis behind our problem definition:
% \begin{itemize}
% \item always see at least one gate
% \item have some state estimation module onboard
% \item no prior knowledge of track configuration -- may change over time
% \item only FPV camera is available as sensor: Let $\pi : I \rightarrow V $ be UAV's policy, where $V = [V_x, V_y, V_z, V_{\psi}] \in \real^4$
% \item define what a feature is, define a policy that operates end-to-end and img to feature, feature to vel
% \end{itemize}

% Important: Define problem in such a way that is generic enough that makes it generalizable to other applications (robotics, gym, etc) besides drone racing

% }

\subsection{Imitation learning for control policy}
\label{subsec:app_imitation}

\noindent{\it Expert trajectory planner and tracker:}
To generate expert data ($\pi^* \big( E \big)$) we use a minimum jerk trajectory planner following the work of \cite{mellinger2011minimum,richter2016polynomial,burri2015real-time} considering a horizon of one gate into the future, and track it using a pure-pursuit path tracking controller.
% The tracking controller minimizes cross track, along track and z-track errors for position and velocity reference set points, obtained by sampling the minimum-jerk trajectory. 
% We integrate the implementation of \cite{burri2015real-time} into AirSim, and 
We generate a dataset of monocular RGB images with their corresponding controller velocity commands.

\noindent{\it Imitation learning algorithm:}
We use behavior cloning (BC), a variant of supervised learning \cite{osa2018algorithmic}, to train the control policy $\pi \big( q(I) \big)$ when minimizing Equation~\ref{eq:main_obj}. We freeze all encoder weights when training the control policy. In total we train 5 policies for the simulation experiments: $BC_{con}$ and $BC_{unc}$, which operate on $z_{con}$ and $z_{unc}$ respectively as features, $BC_{img}$, which uses a pure unsupervised image reconstruction VAE for features, $BC_{reg}$, which uses a purely supervised regressor from image to gate pose as features, and finally $BC_{full}$, which uses a full end-to-end mapping from images to velocities, without an explicit latent feature vector. We train an additional policy $BC_{real}$ for the physical experiments, using unsupervised real-life images along the CM-VAE architecture, as further detailed in Section~\ref{subsec:exp_real}.

To provide a fair comparison between policy classes, we design all architectures to have the same size and structure. $BC$ policies learned on top of representations are quite small, with only $3$ dense layers and roughly $6$K neurons. The end-to-end $BC_{full}$ layout is equivalent to the combination of the Dronet encoder plus $BC$ policy from the other cases, but initially all parameters are untrained.

% \rbnote{show:
% \begin{itemize}
% \item Basic definitions of BC / IL that we're using with learning policy $\pi$
% \item Talk about our separation of feature extraction from policy learning
% \item Show basic derivation for the cross-modal VAE \cite{spurr2018cross} and briefly go over network architectures (img encoder \cite{loquercio2018dronet} / decoders are just simple deconvs and dense layers)
% \item Go over DAGGER training loop?
% \item go over where expert policy comes from -- calculated with min snap trajectory with vel and acc constraints
% \end{itemize}

% }

%% file: inputs/4_exp.tex
% !TEX root = ../root.tex
\section{Results} 
\label{sec:results}
% We detail experimental results on learning representations, along with simulation and physical evaluations of the system. Additional experiments are shown in the supplementary video. 

\input{inputs/4_exp_rep}

\input{inputs/4_exp_sim}
\input{inputs/4_exp_real}

%% file: inputs/4_exp_rep.tex
% !TEX root = ../root.tex

\subsection{Learning Representations}
\label{subsec:exp_rep}

Our first set of experiments aims to valuate the latent representations from three different architectures: 
(i) $q_{reg}$, for direct regression from $I_t \rightarrow z_{reg} = [r, \theta, \phi, \psi] \in \real^4$,
(ii) $q_{unc}$, for the CM-VAE using RGB and pose modalities without constraints: $I_t \rightarrow z_{unc} \in \real^{10}$,
and (iii) $q_{con}$, for the CM-VAE using RGB and pose modalities with constraints: $I_t \rightarrow z_{con} \in \real^{10}$.
% We fixed $z \in \real^{10}$ as the latent space size in all cases except regression.

% \begin{itemize}
% \item ($q_{reg}$) Direct regression: $I_t \rightarrow z_{reg} = [r, \theta, \phi, \psi]$
% \item ($q_{unc}$) Cross-modal VAE using RGB and pose modalities with no special constraints on $z$: $I_t \rightarrow z_{unc}$ (Fig~\rbnote{fig A})
% \item ($q_{con}$) Cross-modal VAE using RGB and pose modalities where the first 4 variables of $z$ are individually decoded into each element of gate poses: $I_t \rightarrow z_{con}$ (Fig~\rbnote{fig B})
% \end{itemize}

% \bsnote{Only data generation is bold-face reader expects a second bold face section.}
We generated 300K pairs of $64\times64$ images along with their corresponding ground-truth relative gate poses using the AirSim simulator \cite{airsim2017fsr}. We randomly sample the distance to gate, aircraft orientation, and gate yaw. $80\%$ of the  data was used to train the network, and $20\%$ was used for validation.

% Our first analysis focuses on the latent space, and the question whether the latent space encodes the information that is useful for our ultimate control task. 
Fig.~\ref{fig:z_vis} displays images decoded over various regions of the latent spaces $z_{con}$ and $z_{unc}$.
 % for both the constrained and unconstrained VAE, as we vary each of the dimensions in the latent space. 
Each row corresponds to variations in $z$ values in one of the $10$ latent dimensions.
We verify that the latent variables can encode relevant information about the gate poses and background information. 
% for both the constraint and the unconstrained cases. 
In addition, the constrained architecture indeed learned to associate the first four dimensions of $z_{con}$ to affect the size, the horizontal offset, the vertical offset and the yaw of the visible gate.
% =======
% Our first analysis focuses on the latent space, and the question whether the latent space encodes the information that is useful for our ultimate control task. Fig.~\ref{fig:z_vis} shows various images that are generated via the decoder part of the network, for both the constrained and unconstrained VAE, as we vary each of the dimensions in the latent space. Each row in the figure corresponds to one of the 10 latent dimensions, and the columns correspond to various values they take. The results highlight that the latent variables are encoding the right amount of information about the gate poses for both the constraint and the unconstrained cases. However, for the constrained case varying the first four dimensions indeed effects the size, the horizontal offset, the vertical offset and the yaw and correctly corresponds to the imposed constraints. 
% >>>>>>> 4a27e568fbadc10531999bb795c41eae1592789b

\begin{figure}[b]
    \centering
    \includegraphics[width=0.25\textwidth]{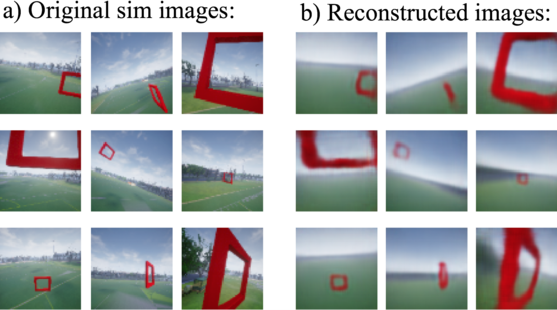}
    \caption{Comparison between original simulated images with their respective CM-VAE reconstructions. Reconstructed images are blurrier than the original, but overall gate and background features can be well represented. 
    % \rbnote{change this figure to show reconstructions from CM-VAE versus regular VAE -- does not capture gates well because they occupy few pixels in the image}
    \fullFigGap }
    \label{fig:recon_unc}
\end{figure}

Fig.~\ref{fig:recon_unc} depicts examples of input images (left) and their corresponding reconstructions  (right). We verify that the reconstruction captures the essence of the scene, preserving both the background and gate pose.
% Explicit consideration of gate pose recovering with the supervised head enforces that the learned weights preserve this information.

\begin{figure}[t]
    \centering
    \includegraphics[width=0.49\textwidth]{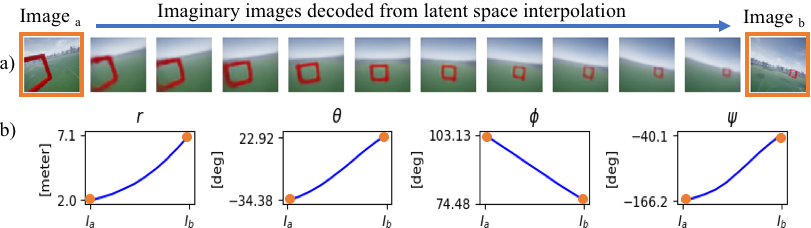}
    \caption{Visualization of latent space interpolation between two simulated images. Smooth interpolation can be perceived in both image and gate pose data modalities. Even background features such as the ground's tilt are smoothly captured.
    }
    \label{fig:interp_sim}
\end{figure}

The smoothness of the latent space manifold with respect to the gate poses and image outputs is a desirable property (\textit{i.e.}, similar latent vectors correspond to similar gate poses).
Intuitively, our single cross-modal latent space should lead to such smooth latent space representation, and our next analysis confirms that such properties emerge automatically. In Fig.~\ref{fig:interp_sim} we show the decoded outputs of a latent space interpolation between the encoded vectors from two very different simulated images. Both images and their decoded poses are smoothly reconstructed along this manifold.

 % two images captured with very different poses. We recover the latent space representations for these two end points and consider an interpolation of the latent vector between them. 
% We then reconstruct both the image as well as the gate poses corresponding to these in-between points and Fig.~\ref{fig:interp_sim} highlight that the 

%We also evaluate the consistency and smoothness of the latent space. In the experiment displayed in Fig~\ref{fig:interp_sim}, we generate two samples in the latent space using images from the dataset: $z_a = q_{con}(I_a)$ and $z_b = q_{con}(I_b)$, and generate imaginary decoded values for both decoded data modalities along a linear interpolation in the latent space. Furthermore, we display variations along individual axes of $z_{unc}$ and $z_{con}$ in Fig~\ref{fig:z_vis}. Note that the addition of constraints in the latent space forces interpretability to the first 4 variables of $z$, which relate directly to physical gate parameters.

Additionally, we quantitatively evaluate the predictions of the three architectures that can recover the gate poses from the images, as shown in Table~\ref{tab:gate_errors}.
% Table~\ref{tab:gate_errors} compares the errors pose test set errors. 
When trained for the same number of epochs, $q_{reg}$, $q_{unc}$, and $q_{con}$ achieve roughly the same error in prediction.
The cross-modal latent space can encode gate pose information slightly better than direct regression, likely due to the additional unsupervised gradient information. Small variations can also be attributed to the training regime using stochastic gradient descent.

% The figure shows that when trained for the same number of epochs until convergence, the cross-modal latent space representations can encode gate pose information better than direct regression, for the same network sizes, likely due to the additional unsupervised gradient information. 
% Figure~\ref{fig:histogram} also shows a detailed histogram of errors in gate pose estimation for the constrained CM-VAE. 
% Overall, the constrained VAE model may be more desirable because in addition to low errors in pose estimation, the latent dimensions are interpretable, making the model more modular and easier to debug. \rbnote{change the point of this paragraph. say that all three networks encode gate poses roughly with the same order of magnitude -- so they are roughly as good as the others}

% \begin{table}[t]
% \caption{Errors for gate pose encoding in different representations}
% \begin{tabular}{l|llllll}
% \label{tab:gate_errors}
%  \textbf{\pbox{1.0cm}{Representation}} & \textbf{\pbox{1.1cm}{Radius \\$r$ [m]}}& \textbf{\pbox{1.3cm}{Azimuth \\$\theta$ [$^\circ$]}} & \textbf{\pbox{.9cm}{Polar $\phi$}}  & \textbf{\pbox{1.4cm}{Yaw $\psi$}} \\
% \hline

% Regression: $z_{reg}$  & $0.41\pm0.42$ & $2.4^\circ\pm4.4$ & $2.5^\circ\pm4.4$  & $11^\circ\pm21$ \\
% CM-VAE: $z_{unc}$  & $0.42\pm0.42$ & \bm{$2.3^\circ\pm4.3$} & \bm{$2.1^\circ\pm3.9$}  & \bm{$9.7^\circ\pm19$} \\
% CM-VAE: $z_{con}$  & \bm{$0.39\pm0.42$} & $2.6^\circ\pm4.2$ & $2.3^\circ\pm4.4$  & $10^\circ\pm19$ \\

% \hline

% \end{tabular}
% \end{table}

\begin{table}[t]
\caption{Average and standard errors for encoding gate poses}
\footnotesize{
\begin{tabular}{m{0.4cm}|cccc}
\label{tab:gate_errors}
 \textbf{$q$} & \textbf{Radius}& \textbf{Azimuth} & \textbf{Polar}  & \textbf{Yaw} \\
  & \textbf{$r$ [m]}& \textbf{$\theta$ [$^\circ$]} & \textbf{$\phi$ [$^\circ$]}  & \textbf{$\psi$ [$^\circ$]}\\
\hline

$q_{reg}$  & $0.41\pm0.013$ & $2.4\pm0.14$ & $2.5\pm0.14$  & $11\pm0.67$ \\
$q_{unc}$  & $0.42\pm0.024$ & \bm{$2.3\pm0.23$} & \bm{$2.1\pm0.23$}  & \bm{$9.7\pm0.75$} \\
$q_{con}$  & \bm{$0.39\pm0.023$} & $2.6\pm0.23$ & $2.3\pm0.25$  & $10\pm0.75$ \\

\end{tabular}
}
\end{table}

% \begin{figure}[t]
%     \centering
%     \includegraphics[width=0.45\textwidth]{figs/histogram}
%     \caption{Histogram showing the density of errors in gate pose decoding for the unconstrained CM-VAE latent representation. Most errors in gate distance are below $0.5$m, and the model learned very precise angles for spherical coordinates $\theta$ and $\phi$. Errors in relative gate yaw are about one order of magnitude higher. \rbnote{probably remove histogram -- doesn't add much}}
%     \label{fig:histogram}
% \end{figure}

%We also evaluate the consistency and smoothness of the latent space. In the experiment displayed in Fig~\ref{fig:interp_sim}, we generate two samples in the latent space using images from the dataset: $z_a = q_{con}(I_a)$ and $z_b = q_{con}(I_b)$, and generate imaginary decoded values for both decoded data modalities along a linear interpolation in the latent space. Furthermore, we display variations along individual axes of $z_{unc}$ and $z_{con}$ in Fig~\ref{fig:z_vis}. Note that the addition of constraints in the latent space forces interpretability to the first 4 variables of $z$, which relate directly to physical gate parameters.

%% file: inputs/4_exp_sim.tex
% !TEX root = ../root.tex

\subsection{Simulated navigation results}
\label{subsec:exp_sim}
Our next set of experiments evaluates control policies learned over five different types of feature extractors. As described in Section~\ref{subsec:app_imitation}, we train behavior cloning policies on top of the CM-VAE latent spaces ($BC_{con}$, $BC_{unc}$), a direct gate pose regressor ($BC_{reg}$), vanilla VAE image reconstruction features ($BC_{img}$), and finally full end-to-end training ($BC_{full}$).

For data collection we generated a nominal circular track with $50$m of length, over which we placed 8 gates with randomized position offsets in XYZ changing at every drone traversal. We collected 17.5K images with their corresponding expert velocity actions while varying the position offset level from 0-3m. $80\%$, or 14K datapoints, were used to train the behavior cloning policies, and the remainder were used for validation.

\begin{figure}[ht]
    \centering
    \includegraphics[width=0.45\textwidth]{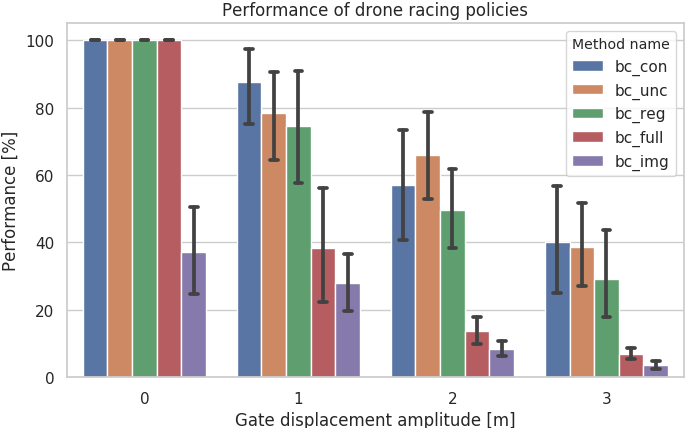}
    \caption{Performance of different navigation policies on simulated track.}
    \label{fig:performance_sim}
\end{figure}

We evaluate our proposed framework under controlled simulation conditions analogous to data collection.
Similarly to previous literature \cite{kaufmann2019beauty,kaufmann2018deep,loquercio2019deep}, we define a success metric of $100\%$ as the UAV traversing all gates in 3 consecutive laps. 
For each data point we average results over 10 trials in different randomized tracks. 
Figure~\ref{fig:performance_sim} shows the performance of different control policies which were trained using different latent representations, under increasing random position offset amplitudes. 
At a condition of zero noise added to the gates, most methods, except for the latent representation that uses pure image compression, can perfectly traverse all gates. As the track difficulty increases, end-to-end behavior cloning performance drops significantly, while methods that use latent representations degrade slower. At a noise level of 3 m over a track with 8 m of nominal radius the proposed cross-modal representation $BC_{con}$ can still achieve approximately $40\%$ success rate, 5X more than end-to-end learning.
We invite the reader to watch the supplementary video for more details.

The three architectures that implicitly or explicitly encode gate positions ($BC_{con}$, $BC_{unc}$, $BC_{reg}$) perform significantly better than the baselines.
This behavior likely spans from the small pixel-wise footprint of gates on the total image, which makes it harder for the vanilla VAE architecture or end-to-end learning to effectively capture the relative object poses.
However, even though the regression features have a relatively good performance in simulation, policy success degrades when exposed to real-world images, as detailed in Subsection~\ref{subsec:exp_real}.

%% file: inputs/4_exp_real.tex
% !TEX root = ../root.tex

\subsection{Real-World Results}
\label{subsec:exp_real}
\begin{wrapfigure}{r}{0.15\textwidth}
    \centering
    \includegraphics[width=0.16\textwidth]{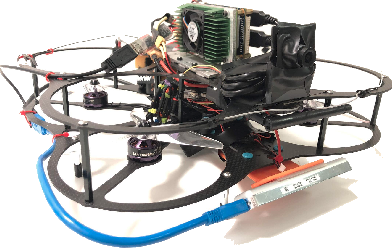}
    \caption{UAV platform.}
    \label{fig:drone}
\end{wrapfigure}

We also validate the ability of the different visuomotor policies to transfer from simulation to real-world deployment.
% cross-modal VAE to combine supervised simulated data and unsupervised real-world images within the same feature extractor, and explore if the behavior cloning policy can be deployed successfully in physical environments. 
Our platform is a modified kit\footnote{https://www.getfpv.com/student-competition-5-bundle.html}, as shown in Figure~\ref{fig:drone}. All processing is done fully onboard with a Nvidia TX2 computer, with 6 CPU cores and an integrated GPU. An off-the-shelf Intel T265 Tracking Camera provides odometry,
 % mounted 45$^{\circ}$ downwards. 
 % The image 
and image processing uses the Tensorflow framework. The image sensor is a USB camera with $83^{\circ}$ horizontal FOV, and we downsize the original images to dimension $128\times72$.

\begin{figure}[t]
    \centering
    \includegraphics[width=0.49\textwidth]{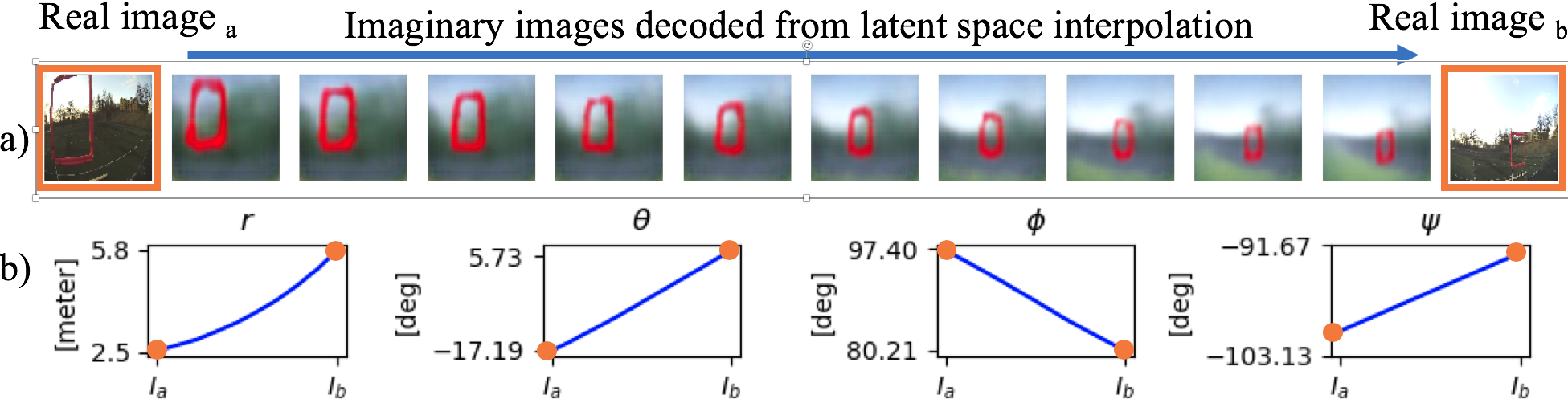}
    \caption{Visualization of smooth latent space interpolation between two real-world images. The ground-truth and predicted distances between camera and gate for images A and B were ($2.0$, $6.0$) and ($2.5$, $5.8$) meters respectively.}
    \label{fig:real_interp}
\end{figure}

First we evaluate how the CM-VAE module, which was learned only with simulated data, performs with real-world images as inputs. We only focus on $z_{con}$ given that it presented the best overall performance in simulation experiments. Fig.~\ref{fig:real_interp} shows that the latent space encoding remains smooth and consistent. 
We train this model with a new simulation dataset composed of $100$K images with size $128\times72$ and FOV equal to our hardware USB camera.

\begin{figure}[t]
    \centering
    \includegraphics[width=0.49\textwidth]{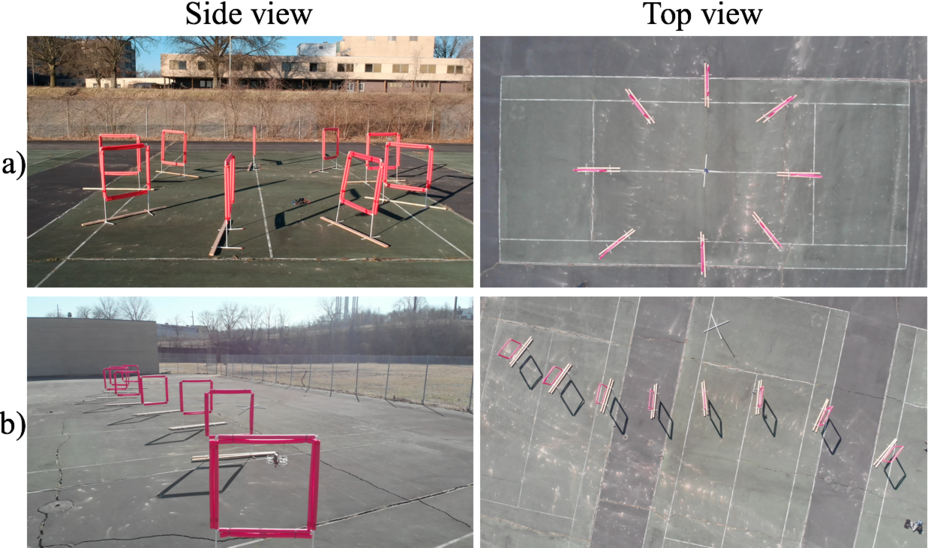}
    \caption{Side and top view of: a) Circuit track, and b) S-shape track.}
    \label{fig:tracks}
\end{figure}

To show the capabilities of our approach on a physical platform, we test the system on a S-shaped track with 8 gates and $45$m of length, and on a circular track with 8 gates and $40$m of length, as shown in Fig~\ref{fig:tracks}.
We compare three policies: $BC_{con}$, $BC_{reg}$, and $BC_{real}$. 
To train this last control policy, $BC_{real}$, we use a CM-VAE trained using not only the $100$K images from simulation, but also additional $20$K unsupervised real-world images. Our goal with this policy is to compare if the use of unsupervised real data can help in the extraction of better features for navigation.
% \rbnote{Furthermore, our proposed framework also has the ability to make use of unsupervised data in the training phase, thereby allowing us to incorporate real-life unlabeled traces into the state representation. }

We display results from both tracks on Table~\ref{tab:real_performance}. The navigation policy using CM-VAE features trained purely in simulation significantly outperforms the baseline policies in both tracks, achieving over 3$\times$ the performance of $BC_{reg}$ in the S-track. The performance gap is even larger on the circuit track, with $BC_{con}$ achieving a maximum of $26$ gates traversed before any collision occurs. It is important to note that some of the circuit trials occurred among wind gusts of up to $20$km/h, a fact that further shows that our policies learned in simulation can operate in physical environments.
 
\begin{table}[t]
\caption{Policy performance in number of gates traversed}
\label{tab:real_performance}
\centering
\begin{tabular}{|l|cc|cc|}
\cline{2-5}
\multicolumn{1}{c|}{} & \multicolumn{2}{c|}{S-Track [12 trials]} & \multicolumn{2}{c|}{Circuit [6 trials]} \\
\multicolumn{1}{c|}{} & Mean & Max & Mean & Max \\
\hline
$BC_{con}$ & \textbf{7.8} & \textbf{8} & \textbf{14.3} & \textbf{26} \\
\hline
$BC_{real}$ & $5.0$ & $7$ & $3.1$ & $5$ \\
\hline
$BC_{reg}$ & $2.3$ & $5$ & $1.2$ & $2$\\
\hline
\end{tabular}
\end{table}

\begin{figure}[t]
    \centering
    \includegraphics[width=0.49\textwidth]{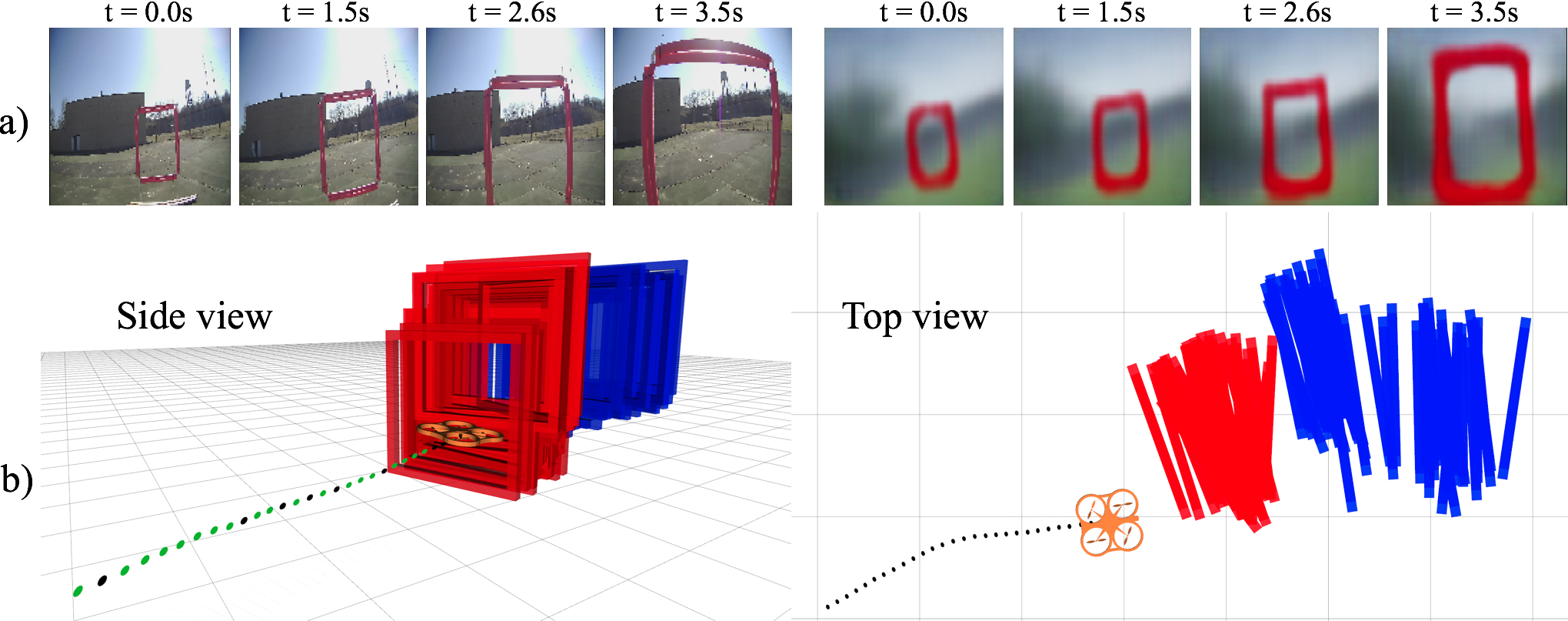}
    \caption{Analysis of a $3$-second flight segment. a) Input images and their corresponding images decoded by the CM-VAE; b) Time history of gate center poses decoded from the CM-VAE (red) and regression (blue). The regression representation has significantly higher offset and noise from the true gate pose, which explains its poor flight performance.}
    \label{fig:rep_comparison}
\end{figure}

We also investigate the cause of the performance drop in $BC_{reg}$ when transferred to real-world data. Table~\ref{tab:gate_estimation} shows the ground-truth and predicted gate distances for different input images. The CM-VAE, despite being trained purely on simulation, can still decode reasonable values for the gate distances. Direct regression, however, presents larger errors. In addition, Figure~\ref{fig:rep_comparison} displays the accumulated gate poses as decoded from both representations during $3$s of a real flight test. The regression poses are noticeably noisier and farther from the gate's true location.

% \begin{table}[t]
% \caption{Examples of distance to gates decoded from real images}
% \footnotesize{
% \centering
% \begin{tabular}{|b{1.1cm}|b{2.2cm}|b{1.7cm}|b{1.9cm}|}
% \hline
% % \begin{tabular}{l|c|c|c}
% \label{tab:gate_estimation}
% \textbf{Image} & \textbf{Ground-truth $[m]$} & \textbf{CM-VAE $[m]$} & \textbf{Regression $[m]$} \\
% \hline
% % index 25 in hand_picked_128_1
% \parbox[l]{0.09cm}{\includegraphics[width=0.05\textwidth]{g_2m}} & \centering{$2.0$} & \centering{\textbf{2.16}} & $\quad\quad4.67$   \\
% \hline
% % index 21 in hand_picked_128_1
% \parbox[l]{0.09cm}{\includegraphics[width=0.05\textwidth]{g_4m}} & \centering{$4.0$} & \centering{\textbf{3.78}} & $\quad\quad5.50$   \\
% \hline
% % index 25 in hand_picked_128_1
% \parbox[l]{0.09cm}{\includegraphics[width=0.05\textwidth]{g_6m}} & \centering{$6.0$} & \centering{\textbf{6.10}} & $\quad\quad6.68$   \\
% \hline
% % index 43 in hand_picked_128_1
% \parbox[l]{0.09cm}{\includegraphics[width=0.05\textwidth]{g_8m}} & \centering{$8.0$} & \centering{\textbf{8.77}} & $\quad\quad9.13$   \\
% \hline
% \end{tabular}
% }
% \end{table}

\begin{table}[t]
\centering
\caption{Examples of distance to gates decoded from real images}
\label{tab:gate_estimation}
\footnotesize{
\begin{tabular}{|m{2.2cm}|c|c|c|c|}
\hline
% \begin{tabular}{l|c|c|c}
\textbf{Image} &
% index 25 in hand_picked_128_1
\includegraphics[width=0.06\textwidth]{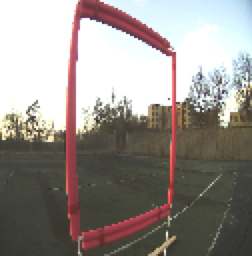} &
% index 21 in hand_picked_128_1
\includegraphics[width=0.06\textwidth]{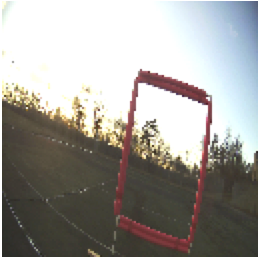} &
% index 25 in hand_picked_128_1
\includegraphics[width=0.06\textwidth]{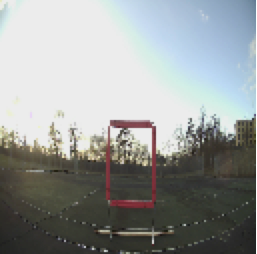} &
% index 43 in hand_picked_128_1
\includegraphics[width=0.06\textwidth]{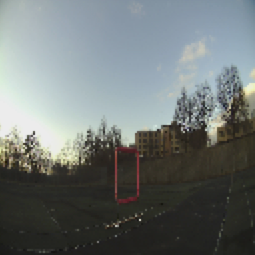} \\
\hline
\textbf{Ground-truth $[m]$} & $2$ & $4$ & $6$ & $8$\\
\hline
\textbf{CM-VAE $[m]$} & \textbf{2.16} & \textbf{3.78} & \textbf{6.10} & \textbf{8.77}\\
\textbf{Regression $[m]$} & $4.67$ & $5.50$ & $6.68$ & $9.13$\\
\hline
\end{tabular}
}
\end{table}

\begin{figure}[t]
    \centering
    \includegraphics[width=0.49\textwidth]{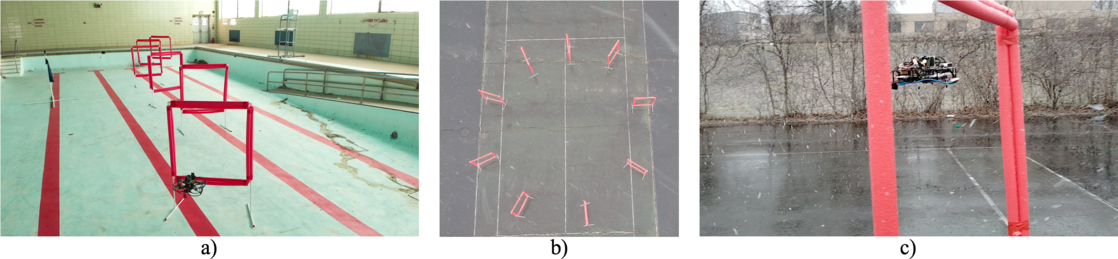}
    \caption{Examples of challenging test environments: a) Indoors, with blue floor with red stripes, and b-c) among heavy snowfall. }
    \label{fig:challenge}
\end{figure}

In the experiments thus far we deployed the learned policies on physical environments that roughly resemble the visual appearances of the simulation dataset. There, all images were generated in a grass field with blue skies, and trees in the background. 
To verify policy and representation robustness to extreme visual changes, we perform additional tests in more challenging scenarios. Fig.~\ref{fig:challenge} shows examples of successful test cases: Fig.~\ref{fig:challenge}a) indoors where the floor is blue with red stripes, and Fig.~\ref{fig:challenge}b-c) among heavy snow. We invite the reader to visualize these experiments in the video attachment (\small{\url{https://youtu.be/VKc3A5HlUU8}})\normalsize.

%% file: inputs/5_discussion.tex
\section{Conclusion and Discussion}
\label{sec:discussion}

In this work we present a framework to solve perception-control tasks that uses simulation-only data to learn rich representations of the environment, which can then be employed by visuomotor policies in real-world aerial navigation tasks. 
% freadily available unsupervised real-world data together with simulated supervised training data to learn robust representations in an efficient manner. 
% \rbnote{change this based on our new key ideas from intro}
At the heart of our approach is a cross-modal Variational Auto-Encoder framework that jointly encodes raw sensor data and useful task-specific state information into a latent representation to be leveraged by a control policy. 
% The framework utilizes the variations introduced by the unsupervised real-world data and the task-specific information via the supervised simulated data and learn robust representations in an efficient manner. 
We provide detailed simulation and real-world experiments that highlight the effectiveness of our framework on the task of FPV drone navigation. Our results show that the use of cross-modal representations significantly improves the real-world performance of control policies in comparison with several baselines such as gate pose regression, unsupervised representations, and end-to-end approaches.

The main finding we can infer from our experiments is that by introducing multiple data modalities into the feature extraction step, we can avoid overfitting to specific characteristics of the incoming data. For example, even though the sizes of the square gates were the same in simulation and physical experiments, their width, color, and even intrinsic camera parameters are not an exact match. 
The multiple streams of information that the CM-VAE encounters regularize its model, leading to better generalization among appearance changes.

From our experiments in Fig.~\ref{fig:performance_sim} we can also infer that features trained for unsupervised image reconstruction can serve as important cues describing the UAV's current state for the control policy, on top of the explicit human-defined supervised parameters.
For example, by using background features such as the line of horizon, a pilot may infer the UAV's current roll angle, which influences the commanded velocities. This remark can serve as an additional motivator for the use of a semi-supervised features extraction module, since it is difficult to hand-define all relevant features for a particular control task.
Another advantage of the CM-VAE architecture is that it can allow the robot operator to gain insights onto the decisions made by the networks. For instance, a human can interpret the decoded gate pose and decoded images in real time and stop the vehicle if perception seems to be abnormal. 
% though only the  because the latent space can be decoded into human-interpretable data such as gate images and gate poses.

Interestingly, $BC_{real}$ did not outperform $BC_{con}$ in our real-world experiments, as we originally expected. 
However, it was still better than $BC_{reg}$. 
We suspect that the drop in performance happens because the dataset used for imitation learning only contained images from simulation, and there is distribution shift in comparison with images used for training the representation. As future work we envision using adversarial techniques such as 
% \cite{zhang2019adversarial,larsen2015autoencoding}
\cite{zhang2019adversarial}
for lowering the distance in latent space between similar scenes encountered in sim and real examples.

Additional future work includes extensions of the cross-modal semi-supervised feature extraction framework to other robotics tasks, considering the use of multiple unsupervised data modalities that span beyond images. We believe that applications such as autonomous driving and robotic manipulation present perception and control scenarios analogous to the aerial navigation task, where multiple modalities of simulated data can be cheaply obtained.

% Additionally, we plan to investigate training more complex control policies on top of the learned latent representations, such as with the use of reinforcement learning (RL). To do such, we can obtain more accurate state representations by applying recurrent dynamics models to the features extracted with the CM-VAE, following techniques discussed in \cite{ha2018world}. Within the context of RL, we would be interested in comparing our architecture with similar concepts such as the use of auxiliary unsupervised tasks \cite{jaderberg2016reinforcement}.

% \rbnote{talk about this paper as well here  -- talk about the challenge of sim2real transfer from an adversarial perspective. say that we envision using extending the VAE work in a similar fashion, training like the VAE-GAN paper \cite{larsen2015autoencoding} }

% \begin{figure}[t]
%     \centering
%     \includegraphics[width=0.49\textwidth]{figs/real_plots}
%     \caption{Visualization of UAV first-person view images along with commanded output velocities. Notice the cyclical patter in velocities, indicative of a 1-gate myopic policy behavior. \rbnote{will remove this fig}}
%     \label{fig:real_plots}
% \end{figure}

% Additionally we are also exploring safety considerations in autonomy that combines the data-driven policies with formal methods based safety reasoning.

% \rbnote{talk about how our approach can generalize well to other domains: describe specific problems and how to apply our techniques}